\documentclass{article}

\usepackage[final]{bdl_2018}

\usepackage[utf8]{inputenc} 
\usepackage[T1]{fontenc}    
\usepackage{hyperref}       
\usepackage{url}            
\usepackage{booktabs}       
\usepackage{amsfonts}       
\usepackage{nicefrac}       
\usepackage{microtype}      
\usepackage{graphicx}
\usepackage{caption}
\usepackage{subcaption}
\usepackage{amsmath}

\usepackage{xcolor}

\title{The Relevance of Bayesian Layer Positioning to Model Uncertainty in Deep Bayesian Active Learning} 

\author{
  Jiaming Zeng\\
  Stanford University\\
  \texttt{jiaming@stanford.edu} \\
  \And
  Adam Lesnikowski \\
  NVIDIA \\
  \texttt{alesnikowski@nvidia.com} \\
  \And
  Jose M. Alvarez\\
  NVIDIA\\
  \texttt{josea@nvidia.com} \\
}

\begin{document}

\maketitle

\begin{abstract}
One of the main challenges of deep learning tools is their inability to capture model uncertainty. While Bayesian deep learning can be used to tackle the problem, Bayesian neural networks often require more time and computational power to train than deterministic networks. Our work explores whether fully Bayesian networks are needed to successfully capture model uncertainty. We vary the number and position of Bayesian layers in a network and compare their performance on active learning with the MNIST dataset. We found that we can fully capture the model uncertainty by using only a few Bayesian layers near the output of the network, combining the advantages of deterministic and Bayesian networks.
\end{abstract}

\section{Introduction}

\begin{figure}[!htb]
    \centering
    \includegraphics[width=0.75\textwidth,trim={0cm 1cm 0 1.1cm},clip]{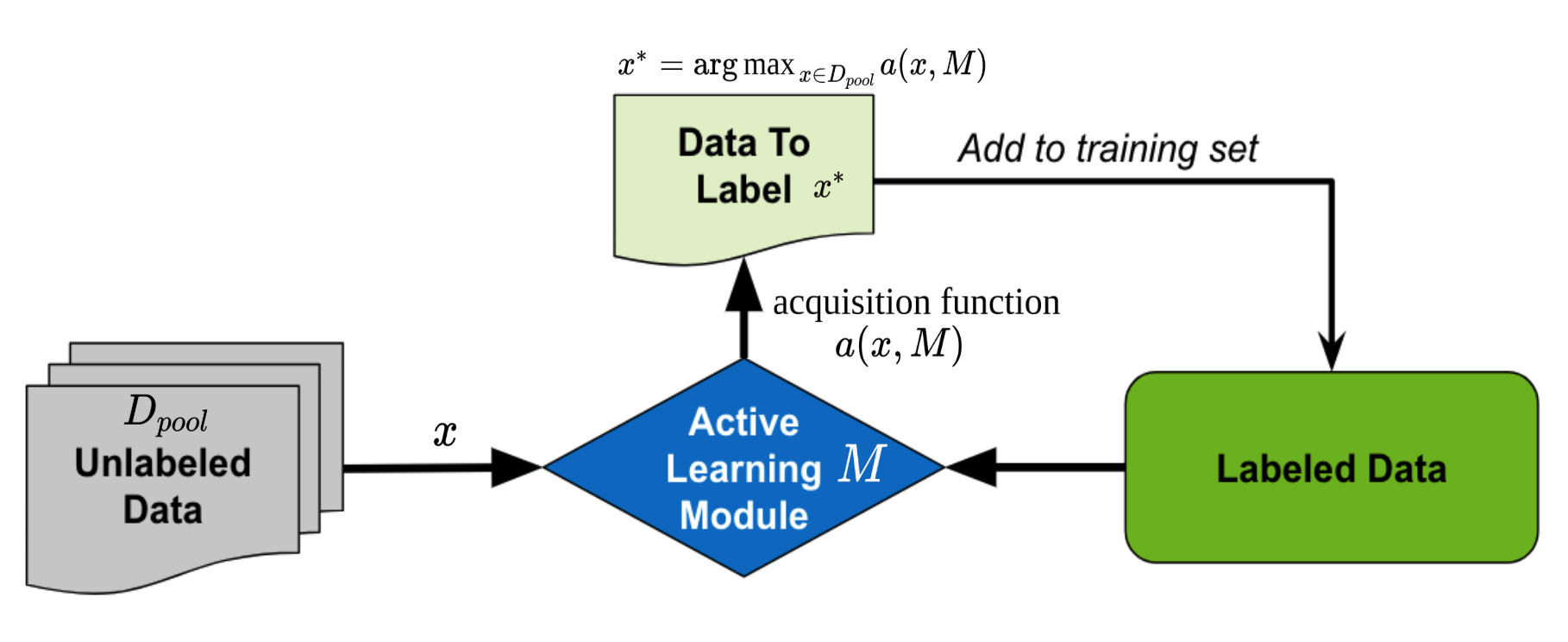}
    \caption{Active learning framework.}
    \label{fig:active-learning}
\end{figure}

Continuously obtaining and labeling data is typically a laborious and costly process necessary for machine learning (ML). Active learning (AL) is a more efficient framework where a system learns from a small amount of data and then chooses which data it would like to label next. In a typical active learning setup, as shown in Figure \ref{fig:active-learning}, an AL module $M$ is trained on a small pool of labeled data at each iteration. An \textit{acquisition function}, often based on the model's uncertainty, uses the model $M$ to select which unlabeled data it would like to ask an external oracle to label. 

While AL is an important technique of machine learning, scaling to high-dimensional data is a major challenge and the existing literature scarce \cite{tong2001active}. \cite{gal2017deep} estimates the network uncertainty through an approximate Bayesian CNN by using Monte Carlo (MC) dropout with convolutional neural networks (CNNs) \cite{gal2015bayesian}. While \cite{gal2017deep} showed significant improvements over existing active learning approaches for high-dimensional data, they note that Bayesian CNNs take a long time to train. Moreover, Bayesian CNNs become increasingly difficult to train in terms of time and complexity as the network size and the number of parameters scale \cite{blundell2015weight, wang2016towards}. 

In this work, we study the relevance of different layers in a Bayesian CNN in capturing the uncertainty of a model. Many work recently have explored how we can extract reliable uncertainty estimates from neural networks \cite{blundell2015weight, hafner2018reliable}; we will focus on a comparison to the methods presented in \cite{gal2017deep}. Instead of MC Dropout, we used Bayesian CNNs with Gaussian approximate variational inference and achieved a comparable level of accuracy as \cite{gal2017deep}. We varied the number and position of Bayesian layers and the weight distribution initialization in our CNNs to examined their ability to capture uncertainty through AL on MNIST. Our results suggest that CNNs with a few Bayesian layers placed near the output can capture the same level of uncertainty as traditional Bayesian CNNs. 

\section{Bayesian Convolutional Neural Networks}

Bayesian CNNs are CNNs with prior probability distributions placed over their model parameters $\mathbf{w}$. Given data $\mathcal{D} = \{\mathbf{X}_i, y_i \}_{i=1}^N$ where $\mathbf{X}_i \in \mathbb{R}^d, y_i \in \mathbb{R}$, we define a prior distribution $p(\mathbf{w}) \sim N(\mu, \Sigma)$ over each parameter. The posterior distribution can be defined as $p(\mathbf{w} | \mathbf{X, y}) = \frac{p(y | \mathbf{X, w})p(\mathbf{w})}{p(y|\mathbf{X})}$. We further define the likelihood or prediction from Bayesian CNNs as 
$p(y^* | x^*, \mathbf{X}, y) = \int p(y^* | \mathbf{x}^*, \mathbf{w}) p(\mathbf{w}|\mathbf{X}, y) d\mathbf{w}$.

Due to the complexity of calculating the model evidence, $p(y|\mathbf{X})$, Bayesian CNNs are typically solved through variational inference (see Appendix \ref{variational_inference}). Instead of calculating the true posterior $p(\mathbf{w} | \mathbf{X, y})$, we approximate it with a simpler distribution $q(\mathbf{w})$. We define the approximate posterior distribution as Gaussian, with $q(\mathbf{w}) \sim N(\mu_q, \sigma_q)$. We note that with CNNs, the Gaussian distribution does not assign negative values to $q(\mathbf{w})$ because the softmax transformed weights are always positive. 

\section{Experimental Setup}
\label{experiments}

We used the same network structure and active learning setup as \cite{gal2017deep} on the MNIST dataset \cite{lecun2010mnist}. We initialized the training with 20 images and acquired 10 images with each cycle, resulting in a total of 1000 images. 

The network architecture is as follows: input-convolution (Conv1) -relu-convolution (Conv2) -relu-max pooling-dropout-dense (Dense1) -relu-dropout-dense (Dense2) -output, with 32 convolution kernels, 4x4 kernel size, 2x2 pooling, dense layer with 128 units, and dropout probabilities 0.25 and 0.5, modeled after the Keras MINST CNN network \cite{chollet2015keras}. 
We experimented with a total of 8 different architectures, as detailed in Table \ref{tab:archs}. By varying the position and number of Bayesian layers, we examined which layers are most important for capturing model uncertainty. 

For acquisition functions, we used: \textit{Random} (baseline): $a(x) =$ unif$(0, 1)$, \textit{Max Entropy} \cite{shannon2001mathematical}: $\mathbb{H}[y|\mathbf{x}, \mathcal{D}] := - \sum_{c} p(y = c | \mathbf{x}, \mathcal{D}) \log p(y = c | \mathbf{x}, \mathcal{D})$, and \textit{Variation Ratios} \cite{freeman1965elementary}: VR$[x] := 1 - \max_y p(y| \mathbf{x}, \mathcal{D})$. 
For each AL acquisition cycle, we trained the network for 200 epochs and used one hundred Monte Carlo samples on the estimated weight posterior, $q(\textbf{w})$, to approximate the probability distribution, $p(y = c| \mathbf{x}, \mathcal{D})$ (see Appendix \ref{estimating-uncertianty}). 

All analysis and implementations are done with Tensorflow Probability (TFP) \cite{dillon2017tensorflow}. To train the Bayesian layers, we used flipout, a more efficient method to decorrelate the gradients within a mini-batch, \cite{wen2018flipout} to derive unbiased stochastic estimates of the gradient. Flipout works by implicitly sampling pseudo-independent weight perturbations for each update in a variational Bayesian network \cite{wen2018flipout}. Experiments are run with the ADAM optimizer \cite{kingma2014adam} with a learning rate of 0.001 and batch size of 64. We set the default initial variance of $q(\mathbf{w})$ to be $\sigma_q \sim N(-3, 0.1)$. In Section \ref{log-variance-mean}, we optimized the performance of Bayesian CNNs by tuning over the initial variance of $q(\mathbf{w})$. Each experiment was repeated and averaged over 3 runs. 




\begin{center}
\begin{table}[!htb]
    \centering
    \footnotesize
    \begin{tabular}{ | l | c | c | c | c | c | c | c | c | c |}
    \hline
     & \textbf{BNN} & \textbf{BNN-1} & \textbf{BNN-2} & \textbf{BNN-3} & \textbf{BNN1} & \textbf{BNN2} & \textbf{BNN3} & \textbf{CNN} \\ 
    \hline
    \textbf{Conv1}  & Bayes & Det & Det & Det & Bayes & Bayes & Bayes & Det \\
    \textbf{Conv2} & Bayes & Det & Det & Bayes & Det & Bayes & Bayes & Det \\
    \textbf{Dense1} & Bayes & Det & Bayes & Bayes & Det & Det & Bayes & Det\\
    \textbf{Dense2} & Bayes & Bayes & Bayes & Bayes & Det & Det & Det & Det \\
    \hline
  \end{tabular}
    \caption{Summary of combinations of Bayesian and deterministic layers in our architectures.}
    \label{tab:archs}
\end{table}
\end{center}

\section{Results}
\label{results}

We compare the results from all eight network architectures in Table \ref{tab:test-errors}. The results with MC Dropout from \cite{gal2017deep} are also included. Due to our use of traditional variational inference, which approximates a more complicated posterior distribution than MC Dropout, the test errors we see are slightly lower than \cite{gal2017deep} but still comparable. 

\begin{figure}[!htb]
    \centering
    \begin{subfigure}{0.45\textwidth}
        \includegraphics[width=\textwidth,trim={0 0 0 1.1cm},clip]{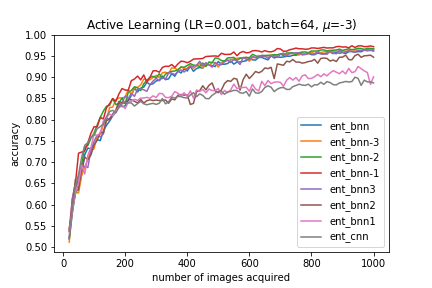}
        \caption{max entropy}
        \label{fig:arch-ent-m-3}
    \end{subfigure}
    \begin{subfigure}{0.45\textwidth}
        \includegraphics[width=\textwidth,trim={0 0 0 1.1cm},clip]{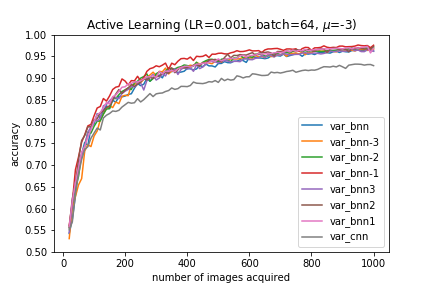}
        \caption{variation ratios}
        \label{fig:arch-var-m-3}
    \end{subfigure}
    \caption{Comparison of the eight experimental architectures for various acquisition functions. (left) plots results for the max entropy acquisition while (right) shows the variation ratios acquisition.}
    \label{fig:experimental-arch}
\end{figure}

Comparing the eight architectures, the ones with no Bayesian layers or with Bayesian layers closer to the input - CNN, BNN1, BNN2, BNN3 - under-performed the fully Bayesian network, BNN. However, the networks with a few Bayesian layers placed near the output - BNN-1, BNN-2, BNN-3 - all outperformed the BNN in capturing uncertainty. In fact, BNN-1 showed the best performance, followed by BNN-2 and then BNN-3. This can be clearly seen in Figure \ref{fig:experimental-arch}. We also note that when using the entire dataset, the deterministic network would outperform the Bayesian networks \cite{blundell2015weight}.

Hence, in AL where less training data is used, we conclude that most of the uncertainty in a model can be captured by using just a Bayesian Dense2 layer. We observe that adding additional Bayesian layers may actually compromise the accuracy without the benefit of added uncertainty modeling. The conclusions observed here are further reiterated in Section \ref{log-variance-mean}, where we studied the effect of capturing uncertainty by varying the Bayesian-ness of the network's prior distribution. 


\begin{center}
\begin{table}[!htb]
    \centering
    \footnotesize
    \begin{tabular}{ | l | c | c | c | c | c | c | c | c | c |}
    \hline
     \textbf{Acquisitions} & \textbf{MC Dropout} & \textbf{BNN} & \textbf{BNN-1} & \textbf{BNN-2} & \textbf{BNN-3} & \textbf{BNN1} & \textbf{BNN2} & \textbf{BNN3} & \textbf{CNN} \\ 
     \hline
     \textbf{Random} & 4.66\% & 6.62\% & \textbf{5.58} \% & 5.71\% & 6.37 \% & 6.62\% & 6.44\% & 6.59\% & 6.90\% \\
     \textbf{Max Ent} & 1.74\% & 3.67\% & \textbf{2.63} \% & 3.22\% & 3.28\% & 7.50\% & 4.62\% & 3.58\% & 10.03\%\\
     \textbf{Var Ratios} & 1.64\% & 3.56\% & \textbf{2.40} \% & 3.00 \% & 3.34\% & 2.70\% & 2.84\% & 3.32\% & 6.48\% \\
    \hline
  \end{tabular}
    \caption{We compared the test error for the 8 architectures with Bayesian CNN to MC Dropout in \cite{gal2017deep}, lower is better. }
    \label{tab:test-errors}
\end{table}
\end{center}

\subsection{Effect of Bayesian-ness of the Prior Initialization}
\label{log-variance-mean}

\begin{figure}[!htb]
    \centering
    \begin{subfigure}{0.32\textwidth}
        \includegraphics[width=\textwidth,trim={0 0 0 1.1cm},clip]{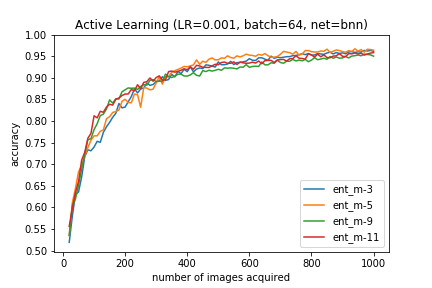}
        \caption{entropy, BNN}
        \label{fig:arch-ent-bnn}
    \end{subfigure}
    \begin{subfigure}{0.32\textwidth}
        \includegraphics[width=\textwidth,trim={0 0 0 1.1cm},clip]{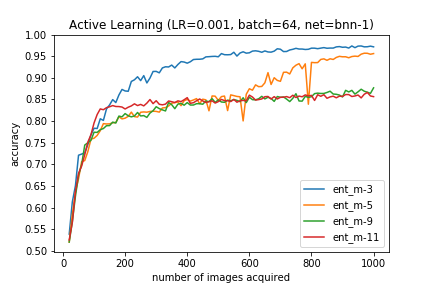}
        \caption{entropy, BNN-1}
        \label{fig:arch-ent-bnn-1}
    \end{subfigure}
    \begin{subfigure}{0.32\textwidth}
        \includegraphics[width=\textwidth,trim={0 0 0 1.1cm},clip]{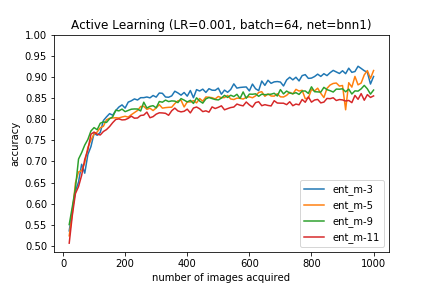}
        \caption{entropy, BNN1}
        \label{fig:arch-ent-bnn1}
    \end{subfigure}
    \begin{subfigure}{0.32\textwidth}
        \includegraphics[width=\textwidth,trim={0 0 0 1.1cm},clip]{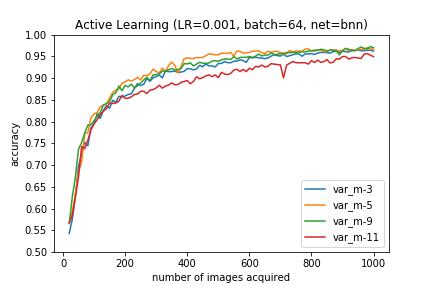}
        \caption{variation ratios, BNN}
        \label{fig:arch-var-bnn}
    \end{subfigure}
    \begin{subfigure}{0.32\textwidth}
        \includegraphics[width=\textwidth,trim={0 0 0 1.1cm},clip]{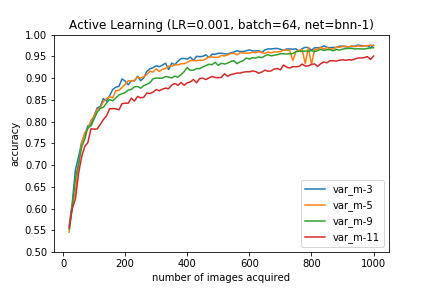}
        \caption{variation ratios, BNN-1}
        \label{fig:arch-var-bnn-1}
    \end{subfigure}
    \begin{subfigure}{0.32\textwidth}
        \includegraphics[width=\textwidth,trim={0 0 0 1.1cm},clip]{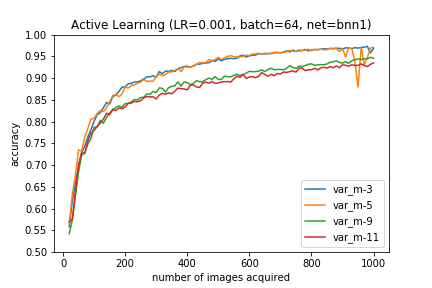}
        \caption{variation ratios, BNN1}
        \label{fig:arch-var-bnn1}
    \end{subfigure}
    \caption{Comparison of different initial variance means for various network architectures.}
    \label{fig:initial_log_variance}
\end{figure}

In addition to the number and position of Bayesian layers, we observe how the prior of the weight distribution, $q(w)$, initialization effects the network's ability to capture uncertainty. We initialize the posterior variance $\sigma_q \sim N(\mu, \sigma)$. To optimize the CNN, we tuned the initial variance mean for $\mu = [-3, -5, -9, -11]$. With softmax transformation, we are essentially defining the initial variance mean as $\ln ({1 + e^{\mu}})$. Hence, lower $\mu$ values initializes the network closer to a deterministic CNN by setting the initial variance of $q(\textbf{w})$ closer to zero and the higher $\mu$ initializes the network to be more Bayesian. We define the \textit{Bayesian-ness} of the initialization by how large the variance mean $\mu$ is. For the architectures in Table \ref{tab:archs}, we selected the fully Bayesian architecture (BNN) and the architectures with one Bayesian layer placed either near the input (BNN1) or the output (BNN-1). In Table \ref{tab:test-errors-best}, we show the test errors using the optimal $\mu$. Compared to Table \ref{tab:test-errors}, there is a noticeable uniform improvement in performance. In Figure \ref{fig:initial_log_variance}, we capture the effect on the different architectures by varying the initial variance mean. 

For max entropy, the fully Bayesian architecture is not affected by the initial $\mu$, as seen in Figure \ref{fig:arch-ent-bnn}. Comparing Figures \ref{fig:arch-ent-bnn-1} and \ref{fig:arch-ent-bnn1}, we note that with only one Bayesian layer, larger initial $\mu$ definitely captures more uncertainty and that Bayesian Dense2 layer is much better than Bayesian Conv1. 

For variation ratios, the Bayesian-ness of the initialization have a greater effect than the network architecture. Comparing Figures \ref{fig:arch-var-bnn-1} and \ref{fig:arch-var-bnn1}, we again note that the larger initial variance performs better and having Bayesian Dense2 layer is better than Bayesian Conv1. 

The results observed here further confirms the conclusions drawn above. The layer most important to capturing uncertainty is Dense2. Moreover, by initializing the Dense2 layer to be more Bayesian, we are able to capture the same level of uncertainty as fully Bayesian CNNs and still maintain the speed and accuracy performance of CNNs.

\begin{center}
\begin{table}[!htb]
    \centering
    \footnotesize
    \begin{tabular}{ | l | c | c | c | c | c | c | c | c | c |}
    \hline
     & \textbf{MC Dropout} & \textbf{BNN} & \textbf{BNN-1} & \textbf{BNN-2} & \textbf{BNN-3} & \textbf{BNN1} & \textbf{BNN2} & \textbf{BNN3} & \textbf{CNN} \\ 
     \hline
     \textbf{Random} & 4.66\% & 6.07\% & \textbf{5.36} \% & 5.71\% & 5.88 \% & 5.93\% & 5.76\% & 6.56\% & 6.90\% \\
     \textbf{Max Ent} & 1.74\% & 3.28\% & \textbf{2.63} \% & 3.15\% & 2.87\% & 7.50\% & 4.62\% & 3.44\% & 10.03\%\\
     \textbf{Var Ratios} & 1.64\% & 2.74\% & \textbf{2.38} \% & 2.69\% & 2.89\% & 2.70\% & 2.59\% & 2.97\% & 6.29\% \\
    \hline
  \end{tabular}
    \caption{We compared the test error for all 8 architectures to the Bayesian CNN with MC Dropout in \cite{gal2017deep}, lower is better. The BNN architectures in Table \ref{tab:archs} are optimized over the initial posterior variance. }
    \label{tab:test-errors-best}
\end{table}
\end{center}

\section{Conclusion}
\label{conclusion}

One major challenge to implementing and using Bayesian CNNs is the time and difficulty required to train them. Our work probes the question of whether fully Bayesian neural networks are needed to effectively capture the uncertainty in a problem. To do this, we experimented with varying the number and position of Bayesian layers for a small CNN. Our results strongly suggest that it is unnecessary to use fully Bayesian CNNs for capturing model uncertainty. We observe that using one or two Bayesian layers (BNN-1, BNN-2) near the output of a network outperforms the fully Bayesian CNN. Moreover, the more Bayesian the layers are, the more uncertainty we can capture. Hence, we can combine deterministic CNNs' accuracy and speed with Bayesian CNNs' ability to capture uncertainty. This would greatly increase the ease of implementation and use of Bayesian deep learning in various applications. 

For future work, we would like to extend the experiment to larger networks, real-world applications, and other methods such as segmentation. Moreover, we would like to extend the experiment to examine the effect of different types of priors on Bayesian CNN performance.

\appendix
\section{Variational Inference}
\label{variational_inference}

Due to the difficulty of calculating the model evidence, $p(y|\mathbf{X})$, Bayesian CNNs are typically solved through approximation methods such as variational inference. Variational inference is a common technique used in statistics to estimate intractable distributions. To solve Bayesian CNNs, we approximate the intractable distribution $p(\mathbf{w} | \mathbf{X, y})$ with a simpler distribution $q(\mathbf{w})$ by minimizing their Kullback-Leibler (KL) divergence. Hence, we solve for  
\[q^*(\mathbf{w}) = \text{arg} \min_{q(\mathbf{w}) \in \mathcal{D}} KL(q(\mathbf{w}) || p(\mathbf{w}|x)) = \text{arg} \min_{q(\mathbf{w}) \in \mathcal{D}} \mathbb{E} \Big( \frac{\log q(\mathbf{w})}{\log p(\mathbf{w}|\mathbf{X})} \Big) \]

Through some algebraic manipulations, we see that minimizing the KL is equivalent to minimizing the negative Evidence Lower Bound (ELBO). We can then solve Bayesian CNNs by using the negative ELBO as loss.
\begin{align*}
    -ELBO &= \mathbb{E} \Big( \frac{\log q(\mathbf{w})}{\log p(\mathbf{w}|\mathbf{X})} \Big) - \log p(\mathbf{X}) \\
    &= \mathbb{E}(\log q(\mathbf{w})) - \mathbb{E} (\log p(\mathbf{w})p(\mathbf{X} | \mathbf{w})) \\
    &= - \mathbb{E}(\log p(\mathbf{X}|\mathbf{w})) + \mathbb{E} \Big( \log \frac{q(\mathbf{w})}{p(\mathbf{w})} \Big)
\end{align*}

\section{Estimating Model Uncertainty}
\label{estimating-uncertianty}
Using variational inference, we approximate the true posterior with an simpler distribution . Then, we can estimate the uncertainty of each prediction by marginalizing over the approximate posterior using Monte Carlo integration. 
\begin{align*}
    p(y = c| \mathbf{x}, \mathcal{D}) &= \int p(y = c| \mathbf{x}, \mathbf{w}) p(\mathbf{w}| \mathbf{x}, \mathcal{D}) d\mathbf{w} \\
    &\approx \int p(y = c| \mathbf{x}, \mathbf{w}) q^*(\mathbf{w}) d\mathbf{w} \\
    &\approx \frac{1}{T} \sum_{t=1}^T p(y = c|\mathbf{x}, \mathbf{\hat{w}}_t)
\end{align*}
We perform $T$ forward passes through the Bayesian CNN, each time sampling a different set of weights. The prediction results from all $T$ passes are then averaged together to give us the approximate predictive distribution. 





\bibliographystyle{unsrt}
\bibliography{nips_bdl}

\end{document}